\documentclass[journal]{IEEEtran}
\makeatletter\if@twocolumn\PassOptionsToPackage{switch}{lineno}\else\fi\makeatother

\usepackage{graphicx}
\usepackage[T1]{fontenc}
\ifCLASSINFOpdf
\else
\fi
\usepackage{amsmath}
\usepackage[cmintegrals]{newtxmath}
\usepackage{url,multirow,morefloats,floatflt,cancel,tfrupee}
\makeatletter

\AtBeginDocument{\@ifpackageloaded{textcomp}{}{\usepackage{textcomp}}}
\makeatother
\usepackage{colortbl}
\usepackage{xcolor}
\usepackage{pifont}
\usepackage[nointegrals]{wasysym}
\makeatletter

\def\mcWidth#1{\csname TY@F#1\endcsname+\tabcolsep}

\def\cAlignHack{\rightskip\@flushglue\leftskip\@flushglue\parindent\z@\parfillskip\z@skip}
\def\rAlignHack{\rightskip\z@skip\leftskip\@flushglue \parindent\z@\parfillskip\z@skip}

\@ifundefined{etal}{}{}

\usepackage{ifxetex}
\ifxetex\else\if@twocolumn\@ifpackageloaded{stfloats}{}{\usepackage{dblfloatfix}}\fi\fi

\AtBeginDocument{
\expandafter\ifx\csname eqalign\endcsname\relax
\def\eqalign#1{\null\vcenter{\def\\{\cr}\openup\jot\m@th
  \ialign{\strut$\displaystyle{##}$\hfil&$\displaystyle{{}##}$\hfil
      \crcr#1\crcr}}\,}
\fi
}

\AtBeginDocument{%
  \@ifpackageloaded{endfloat}%
   {\renewcommand\efloat@iwrite[1]{\immediate\expandafter\protected@write\csname efloat@post#1\endcsname{}}}{\newif\ifefloat@tables}%
}%

\def\BreakURLText#1{\@tfor\brk@tempa:=#1\do{\brk@tempa\hskip0pt}}
\let\lt=<
\let\gt=>
\def\processVert{\ifmmode|\else\textbar\fi}

\@ifundefined{subparagraph}{
\def\subparagraph{\@startsection{paragraph}{5}{2\parindent}{0ex plus 0.1ex minus 0.1ex}%
{0ex}{\normalfont\small\itshape}}%
}{}

\newcommand\role[1]{\unskip}
\newcommand\aucollab[1]{\unskip}
  
\@ifundefined{tsGraphicsScaleX}{\gdef\tsGraphicsScaleX{1}}{}
\@ifundefined{tsGraphicsScaleY}{\gdef\tsGraphicsScaleY{.9}}{}
\def\checkGraphicsWidth{\ifdim\Gin@nat@width>\linewidth
	\tsGraphicsScaleX\linewidth\else\Gin@nat@width\fi}

\def\checkGraphicsHeight{\ifdim\Gin@nat@height>.9\textheight
	\tsGraphicsScaleY\textheight\else\Gin@nat@height\fi}

\def\fixFloatSize#1{}
\let\ts@includegraphics\includegraphics

\def\inlinegraphic[#1]#2{{\edef\@tempa{#1}\edef\baseline@shift{\ifx\@tempa\@empty0\else#1\fi}\edef\tempZ{\the\numexpr(\numexpr(\baseline@shift*\f@size/100))}\protect\raisebox{\tempZ pt}{\ts@includegraphics{#2}}}}

\AtBeginDocument{\def\includegraphics{\@ifnextchar[{\ts@includegraphics}{\ts@includegraphics[width=\checkGraphicsWidth,height=\checkGraphicsHeight,keepaspectratio]}}}

\DeclareMathAlphabet{\mathpzc}{OT1}{pzc}{m}{it}

\def\URL#1#2{\@ifundefined{href}{#2}{\href{#1}{#2}}}

\def\UrlOrds{\do\*\do\-\do\~\do\'\do\"\do\-}%
\g@addto@macro{\UrlBreaks}{\UrlOrds}

\edef\fntEncoding{\f@encoding}

\makeatother

\newif\ifmultipleabstract\multipleabstractfalse%
\usepackage{tabulary}
\makeatletter
\AtBeginDocument{\@ifpackageloaded{longtable}{%
\def\LT@makecaption#1#2#3{%
  \LT@mcol\LT@cols c{\hbox to\z@{\hss\parbox[t]\LTcapwidth{%
    \sbox\@tempboxa{#1{#2: } #3}%
    \ifdim\wd\@tempboxa>\hsize
      #1{#2: }\textsc{#3}%
    \else
      \hbox to\hsize{\hfil\box\@tempboxa\hfil}%
    \fi
    \endgraf\vskip\baselineskip}%
  \hss}}}
}{}}
\makeatother

   \makeatletter
  \def\fig@textbf{\textbf}
   \AtBeginDocument{\renewcommand\floatc@plain[2]{\setbox\@tempboxa\hbox{{\footnotesize#1.}\footnotesize\hskip.5em#2}%
    \ifdim\wd\@tempboxa>\hsize {\fig@textbf{\footnotesize#1.}}\footnotesize\hskip.5em#2\par
        \else\hbox to\hsize{\hfil\box\@tempboxa\hfil}\fi}}
    \makeatother
  
\usepackage{float}

\begin{document}

%

        \title{Modeling Generalized Specialist Approach To Train Quality Resilient Snapshot Ensemble}
      
\author{Ghalib~Ahmed~Tahir ~\IEEEmembership{Member,~IEEE,}, ~Chu~Kiong~Loo ~\IEEEmembership{Senior Member,~IEEE,} and Zongying Liu ~\IEEEmembership{Member,~IEEE,}}

\maketitle 

%
\IEEEpeerreviewmaketitle

\section{Abstract}
Convolutional neural networks (CNNs) apply well with food image recognition due to the ability to learn discriminative visual features. Nevertheless, recognizing distorted images is challenging for existing CNNs. Hence, the study modelled a generalized specialist approach to train a quality resilient ensemble. The approach aids the models in the ensemble framework retain general skills of recognizing clean images and shallow skills of classifying noisy images with one deep expertise area on a particular distortion. Subsequently, a novel data augmentation random quality mixup (RQMixUp) is combined with snapshot ensembling to train G-Specialist. During each training cycle of G-Specialist, a model is fine-tuned on the synthetic images generated by RQMixup, intermixing clean and distorted images of a particular distortion at a randomly chosen level. Resultantly, each snapshot in the ensemble gained expertise on several distortion levels, with shallow skills on other quality distortions. Next, the filter outputs from diverse experts were fused for higher accuracy. The learning process has no additional cost due to a single training process to train experts, compatible with a wide range of supervised CNNs for transfer learning. Finally, the experimental analysis on three real-world food and a Malaysian food database showed significant improvement for distorted images with competitive classification performance on pristine food images. \mbox{}\protect\newline 
    \begin{IEEEkeywords}Deep learning, Ensemble learning, Data Augmentation, Malaysian Food database\end{IEEEkeywords}
\section{INTRODUCTION AND CURRENT METHODOLOGIES}
\begin{figure}[h]
\centering
\includegraphics[width=3.3in]{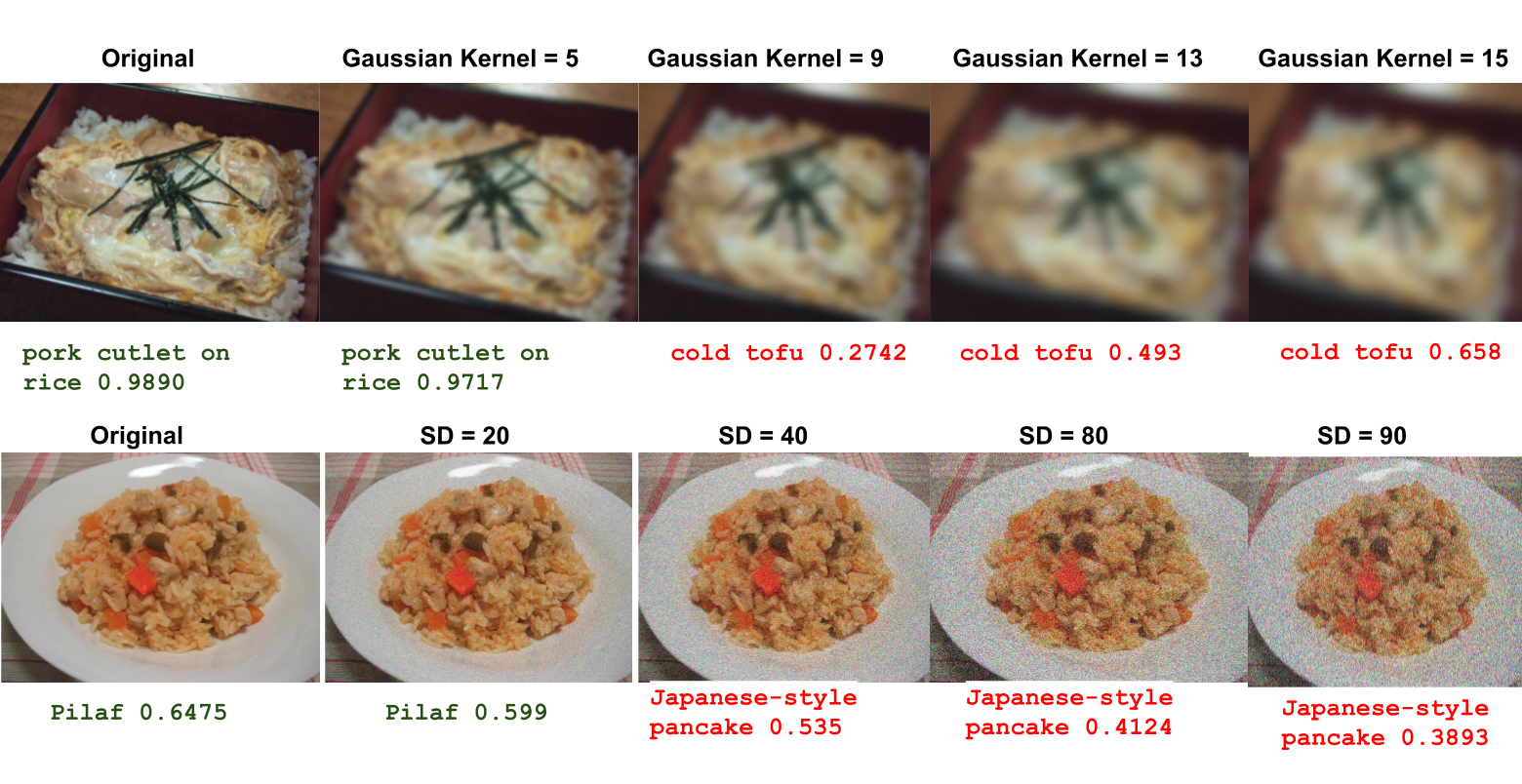}
\caption{Effects of image quality on CNNs}
\label{effectofimagequality}
\end{figure}

A vital aspect overlooked in existing methodology framework is the visual quality of food images. In real-world computer vision applications, images undergo various distortions, such as blur or additive noise during capturing or transmission. Generally, current CNNs shows extreme sensitivity to these image distortions \cite{4}\cite{5}\cite{6}\cite{9}. Nevertheless, Dodge and Karam \cite{2} stated that the distorted images do not represent adversarial samples for CNNs but the images cause a significant reduction in classification performance. Figure \ref{effectofimagequality} shows the effect of image quality on the prediction performance of CNNs trained on pristine food images. When the predicted label is correct, its confidence decreases significantly with the increasing high distortion severity, indicating that features learnt from the dataset of pristine images are not invariant to image distortion nor useable for applications with varying image quality. \mbox{}\protect\newline Past studies addressed this challenge by fine-tuning with corrupted images at all distortion levels \cite{6}\cite{3}, making the CNN model resilient to quality distortions.
Nonetheless, the method requires large distorted datasets at all quality levels, which are not available and is time-consuming. Besides, fine-tuning with low-quality images decreases the accuracy of the clean food photos and increases the epochs to converge the model. Although other methods feature quantization \cite{8}], stability learning \cite{7}, and Deep correct \cite{1} reduce the impact of image quality on CNNs, requiring architectural level changes in the pre-trained models. Similarly, stability learning and feature quantization decrease the Top 1 accuracy, whereas Deep correct increases the floating-point operations (FLOPS) during training and testing.
\par The study extended the CNN-based approach by combining multiple networks with a robust ensemble framework G-Specialist resilient against large variations in quality distortions. Unlike image denoising methods, such as Block Matching 3D denoising (BM3D) \cite{16} and Non-locally Centralized Sparse Representation (NCSR) \cite{17} (which predict the known distortion intensity during training and testing), the G-Specialist approach trains an ensemble for all distortion levels simultaneously. 
\par Accordingly, the study introduced a generalized specialist approach in deep learning with each CNN in the ensemble carrying the general skills of recognizing pristine images, specialist skills on a particular distortion (Gaussian noise or blur) at all distortion levels, and shallow skills on other distortions. Thus, a suitable solution to train generalized specialists during the ensemble learning process is data augmentation. 
\par The proposed RQMixup increases the quantifiable data value without increasing the size of training data or change in the model architecture. Moreover, the approach generates a synthetic sample coupled with snapshot ensembling [10], creating a mixture of experts resilient to distortions and improves classification performance on pristine images without extra training time or increasing the training dataset.
\par The study has done experimentation on Malaysian Food database[ref], the largest database yet to the authors’ best knowledge and experimented on other three publicly available large food databases that will be publicly available online [25].
    
\section{METHOD}
\subsection{General Architecture} 
\begin{figure}[h]
\centering
\includegraphics[width=3.3in]{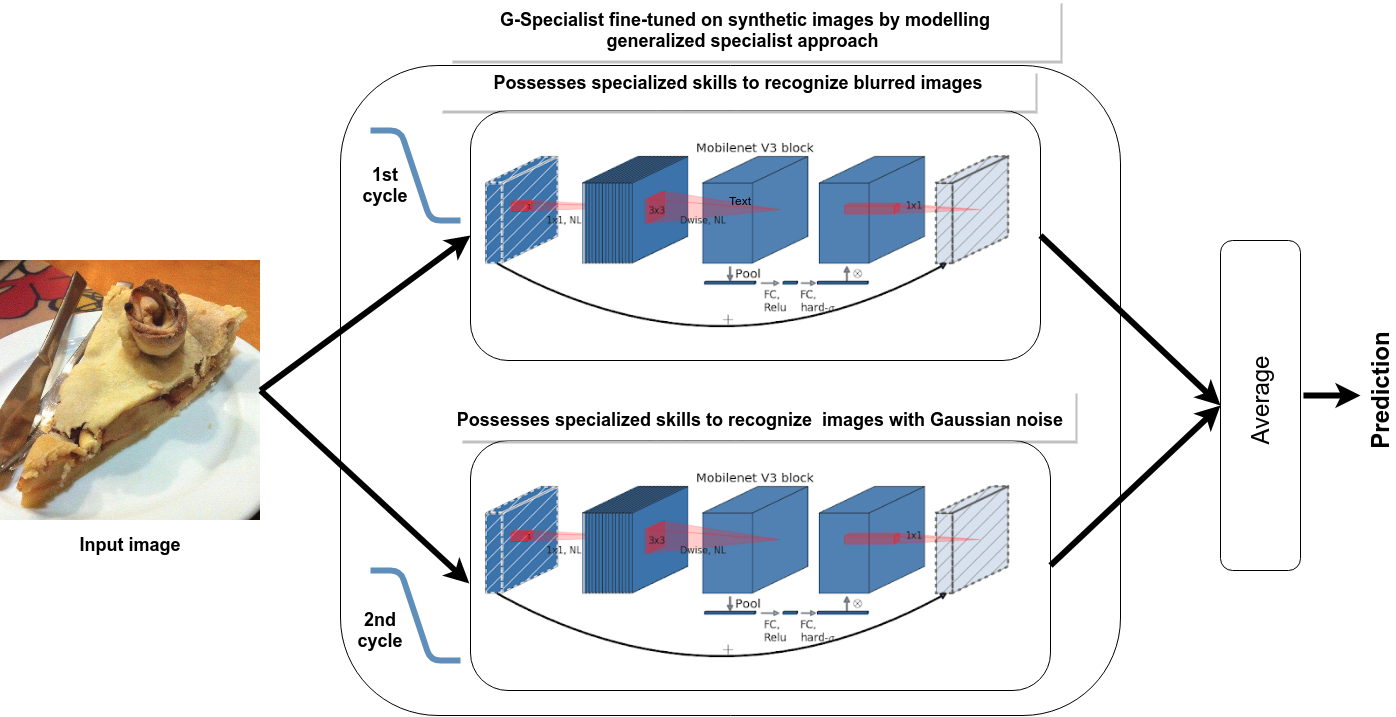}
\caption{The proposed quality resilient ensemble for image recognition by modeling generalized specialist approach during training}
\label{general_architecture}
\end{figure}

\mbox{}\protect\newline The proposed framework is based on the recent emergence of efficient neural networks \cite{18}\cite{19}\cite{20}. Furthermore, the study  selected MobileNetV3 \cite{18} as it is resource-efficient, a crucial requirement of on-device computation for edge devices. Next, RQMixup was combined with snapshot ensembling to train a mixture of robust experts G-Specialist that are resilient to extreme image distortions. Figure \ref{general_architecture} shows the general architecture of the framework. 

\subsection{Random MixUp Data Augmentation}
Data involves driving and powering artificial intelligence (equivalent to fuel for learning algorithms), a fundamental ingredient in supervised machine learning problems with a quantified value in algorithmic classification tasks measurable by the Data Shapley method \cite{11} for supervised classification tasks. Hence, the study developed a generic method that increases the quantified value of data sets in algorithmic decisions to train robust models resilient to different quality distortions without significant changes at the architecture level. 
\par
\begin{figure}[h]
\centering
\includegraphics[width=3.3in]{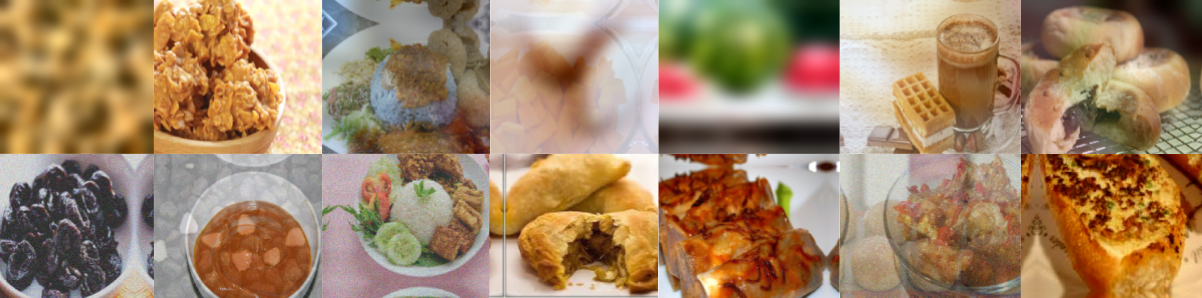}
\caption{Samples of synthetic images to fine-tune a model in a single cycle.}
\label{synthethicimage}
\end{figure}

 Accordingly, the study chose RQMixup that generates a synthetic image to fine-tune quality resilient deep learning models and used two copies of the same training dataset, applying randomly chosen distortion levels to one copy of the dataset during training. Next, raw vectors of cleaned and noisy images and their corresponding labels were mixed using a strategy similar to mixup. For example, raw vectors of a clean set of images in each batch are mixed by applying blurring on the copy of the same set of the training dataset for the Gaussian blur, randomly selecting the blur intensity (1, 3, 5, 7, 9, 11, 13, and 15) for each batch. Consequently, fine-tuning the model with synthetic images turns the model into an expert on all intensity levels of Gaussian blur in the test samples. Equations 1 and 2 compute the mixup of feature-target vectors of the pristine and corrupted images.
\let\saveeqnno\theequation
\let\savefrac\frac
\def\dispfrac{\displaystyle\savefrac}
\begin{eqnarray}
\let\frac\dispfrac
\gdef\theequation{1}
\let\theHequation\theequation
\label{dfg-6402a1a1e0bc}
\begin{array}{@{}l}x_f=\;\lambda x_i+(1-\lambda)x_n\end{array}
\end{eqnarray}
\global\let\theequation\saveeqnno
\addtocounter{equation}{-1}\ignorespaces 
\vskip-1.5\baselineskip 
\let\saveeqnno\theequation
\let\savefrac\frac
\def\dispfrac{\displaystyle\savefrac}
\begin{eqnarray}
\let\frac\dispfrac
\gdef\theequation{2}
\let\theHequation\theequation
\label{dfg-b3bda47d2d69}
\begin{array}{@{}l}y_f=\;\lambda y_i+(1-\lambda)y_n\end{array}
\end{eqnarray}
\global\let\theequation\saveeqnno
\addtocounter{equation}{-1}\ignorespaces 
Specifically, $x_n $ is the raw input vector and $y_n $ is the one-hot label encodings of the noisy image at the random distortion level, $x_i $ is the raw input vector, and $y_i $ is the one-hot label encodings of the pristine image. Meanwhile, Figure \ref{synthethicimage} shows samples of synthetic images generated to fine-tune a model in a single cycle. Notably, the proposed method does not increase the training images required to fine-tune the model

\subsection{Snapshot Ensembling to Train G-Specialist by Modeling Generalized Specialist Approach}Although the single model fine-tuned on synthetic images generated by the RQMixup delivers all the distortion levels of particular noise, multiple types of noises exist, such as Gaussian additive noise and blur. Therefore, the study employed snapshot ensembling to train generalized specialists on various distortions while inheriting the core benefit Train 1 Get M experts. 

The process follows Loshchilov et al.’s \cite{12} cyclic annealing schedule to converge with multiple local minima and train M generalized specialists in a single training process using synthetic images generated by RQMixup.  The learning rate was lowered at a very fast pace in each cycle, converging the model towards local minima after a minimum of 32 epochs. The process was repeated multiple times to obtain M generalized experts through multiple convergences. The learning rate alpha is as follows:
\begin{eqnarray*}\alpha(t)\;=\;f(mod(t-1,\left[T/M\right])) \end{eqnarray*}
Specifically, T is the total number of iterations and f is a monotonically decreasing function. Thus, the training process of the whole framework was split into M cycles whereby each cycle starts with a higher learning rate annealed towards a smaller learning rate to converge towards local minima. A larger learning rate at the start of the cycle provides the model energy to escape a critical point while a smaller learning rate converges the model  to local minima. In the experiment, f is shifted cosine function following Loshchilov et al. \cite{12}
\begin{eqnarray*}\alpha(t)\;=\frac{\alpha_0}2\left(\cos\;\left({\textstyle\frac{\pi\;mod(t-1,\;\left[T/M\right])}{\left[T/M\right]}}\right)+1\right) \end{eqnarray*}
The M-value is 2 for the Gaussian noise and blur. The total cycles were kept equivalent to the number of distortions because during each process cycle, the generalized specialist was trained on a particular distortion by applying snapshot ensembling. The method enabled each snapshot to classify the clean images  with a particular distortion, such as Gaussian additive noise at all intensity levels with confidence.
Although the method  is not an expert on another distortion, such as blurring, it possesses shallow skills of recognizing blur images. Hence, the data augmentation in the second cycle mixes the clean images with a randomly chosen intensity level of blurring, turning the second snapshot of the model to an expert on all distortion levels in the images with shallow skills on other distortions.  \mbox{}\protect\newline As the model usually converges to local minima at the end of every cycle, a snapshot of the model weights was taken before training the next generalized experts on another distortion type. Furthermore, an M snapshot was used in the final ensemble after training the model for M cycles. Besides making each snapshot expert to the particular type of noise, the method also inherits the benefits of ensembling for clean photos as each snapshot converges to different local minima. Contrary to a single model, the ensemble of these snapshots has lower test errors, as shown in Table \ref{tw-2} and \ref{tw-3}. 
\mbox{}\protect\newline The ensemble prediction during testing is the average of M experts equivalent to the total snapshots. Let I (w, h, and c) represent a pre-processed input image of size w × h pixels, and c is the number of channels of the input image. After feeding the image to the input conv2d of a snapshot in the ensemble, g(x) will be the softmax score from a snapshot. Additionally, the final output of the ensemble is an average of all models.
\begin{eqnarray*}g_{ensemble}=\;\frac1M\sum_0^{M-1}g(x) \end{eqnarray*}

\section{Experiment And Results}
The experiments were performed  on a server with 25 GB  of random access memory (RAM) equipped with 16 GB of the graphics processing unit (GPU). Besides, all the experiments were performed using Python on three publically available large food databases (FOOD101 \cite{13}, UECFOOD100 \cite{14}, UECFOOD256 \cite{15}) and a database of Malaysian foods. Table I summarizes the crucial attributes of the datasets.
\begin{table}[!htbp]
\caption{{FOOD RECOGNITION DATASETS} }
\label{tw-163c53bbe5de}
\def\arraystretch{1}
\ignorespaces 
\centering 
\begin{tabulary}{\linewidth}{LLLL}
\hline Dataset & Classes & Instances & Training/Testing\\
\hline 
FOOD101 &
  101 &
  101330 &
  75,750/25,250\\
UECFOOD100 &
  100 &
  14361 &
  12864/1497\\
UECFOOD256 &
  256 &
  31148 &
  28033/3115\\
Malaysian Food &
  775 &
  37198 &
  33724/3474\\
\hline 
\end{tabulary}\par 
\end{table}
\begin{figure}[h]
\centering
\includegraphics[width=3.3in]{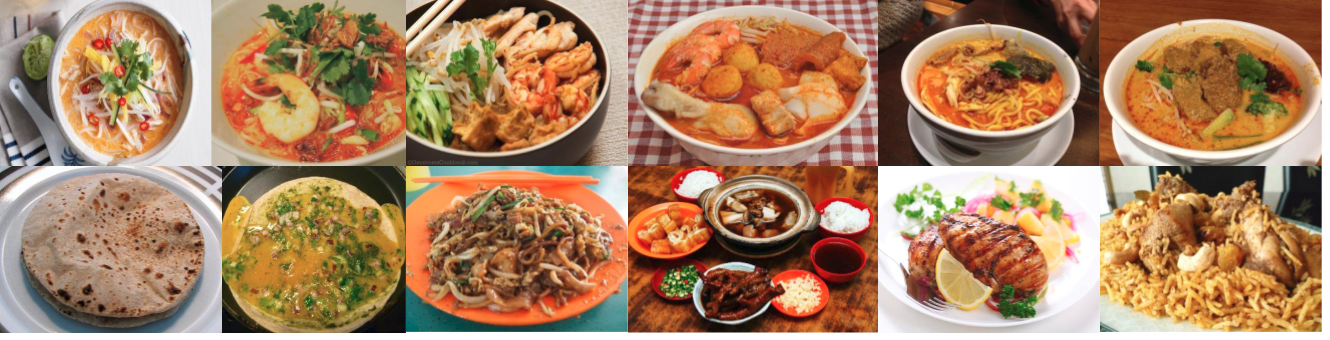}
\caption{The top row are the same class of Malaysian food while
the bottom row is random images of various classes of Malaysian food.}
\label{random_vs_gwr}
\end{figure}

\subsection{G-Specialist Analysis on Distorted Test Images}
The G-Specialist was evaluated against the ensemble model on test images distorted with the Gaussian noise and blurring, with a standard deviation of 10 to 100. Meanwhile, the Gaussian kernel during blurring of test images was varied from 1 to 15. Figure \ref{f-849f6339b7c7} shows the test image results distorted with Gaussian noise and blur for all datasets, denoting that the G-Specialist significantly outperformed the ensemble fine-tuned on pristine images with a significant difference for a higher noise level. Additionally, both models had similar training costs.
\par
\begin{figure}[h]
\centering
\includegraphics[width=3.3in]{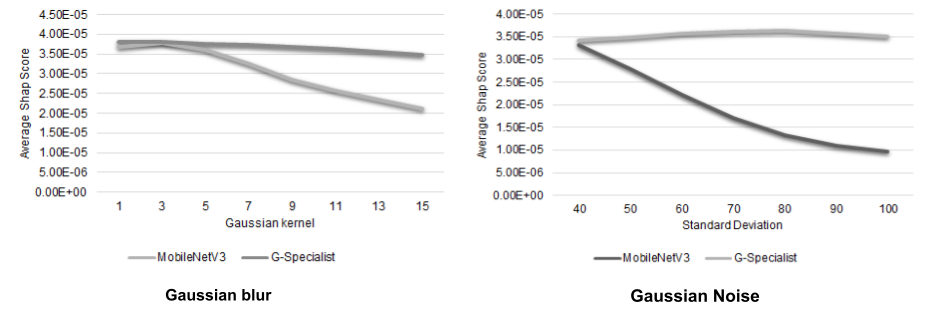}
\caption{The average Shapley value shows that as the noise level increases, the average Shapley value on noisy images decreases}
\label{shapscore}
\end{figure}

 The study also identified whether synthetic data adds value to the model by employing a gradient explainer \cite{21} to compute average Shapley values on test images distorted with Gaussian noise and blur. Figure \ref{shapscore} shows that the noise level increases while the test images average Shapley value decreases . Nonetheless, the Shapley values for G-Specialists were significantly higher than the ensemble fine-tuned with pristine images, suggesting improved robustness by increasing the data value
\bgroup
\fixFloatSize{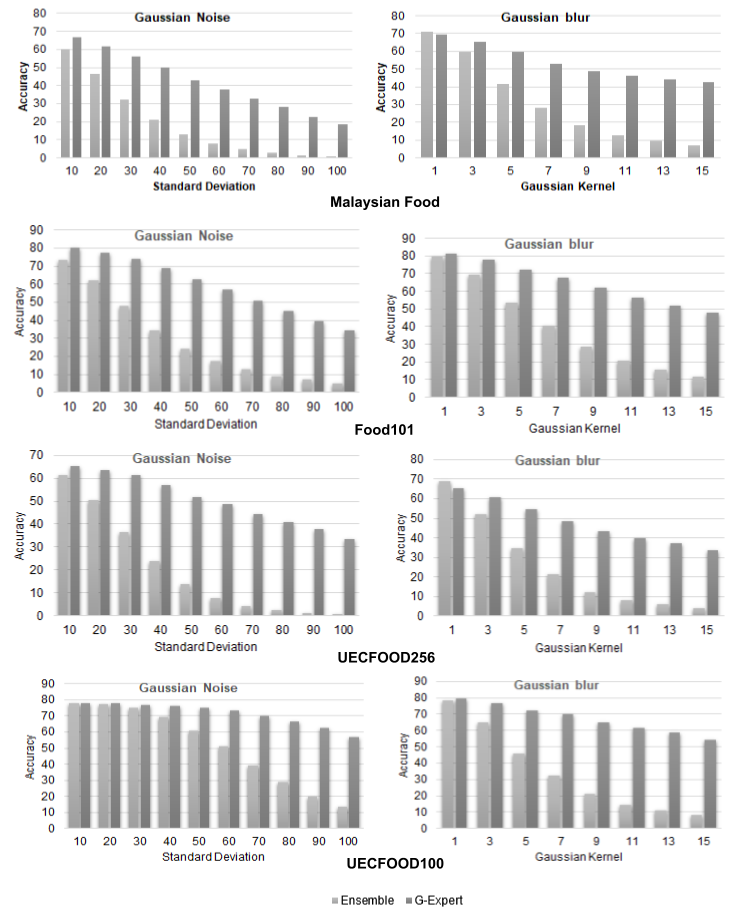}
\begin{figure}[!htbp]
\centering \makeatletter\IfFileExists{Gspecialist.png}{\includegraphics{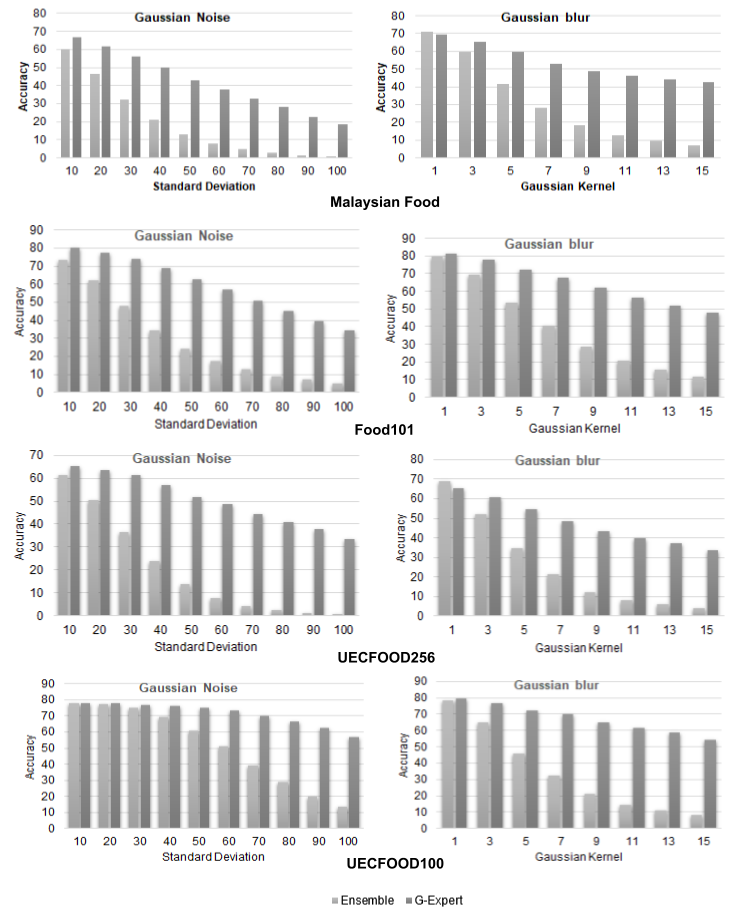}}{}
\makeatother 
\caption{{The accuracy (\%) on distorted data for Food101, UECFOOD256, UECFOOD101, Malaysian Food and comparison with an ensemble of Mo-
bileNetV3 after fine-tuning
}}
\label{f-849f6339b7c7}
\end{figure}
\egroup

\subsection{Results on Pristine Test Images}
The results in Table \ref{tw-2} of Top 1 and Top 3 accuracy show that G-Specialist outperformed all other variants of MobileNet including the single model of MobileNet-V3 on the pristine test data. The study only reported the highest performers, with more results seen in [24], [26], [27], [25], and [28].
 
\subsection{Results on Malaysian Food Database}
\par
\begin{figure}[!ht]
\includegraphics[width=3.3in]{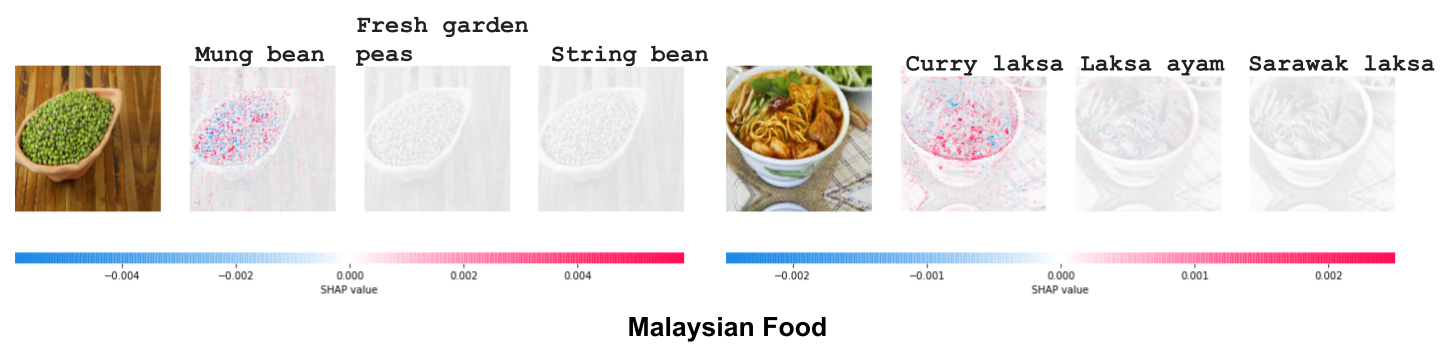}
\caption{Red pixels show the SHAP values of the image positively contribute towards the class, whereas blue pixels show the pixels that reduce the probability of class. The top 3 outputs are explained with ranked\_outputs = 3}
\label{visualization}
\end{figure}

 Malaysian Food database is comprised of 775 classes with 37,198 instances and these classes were selected based on recommendations from an expert dietician. The images were collected from Google search, Bing search, social media and authors’ data using smartphones for capturing food images. Besides, the study used the same experimental settings as the other three datasets. Figure \ref{f-849f6339b7c7} depicts that G-Specialist outperformed other methods on test images distorted with Gaussian noise and blur. Meanwhile, Table \ref{tw-3} suggests that the G-Specialist was better at recognizing food images than a single model and other variants of MobileNet. Additionally, Figure \ref{visualization} illustrates that the study approach in considering relevant pixels for recognizing food images increases the model transparency.

\begin{table}[!htbp]
\caption{ACCURACY (\%) ON PRISTINE DATA FOR FOOD101, UECFOOD256, UECFOOD101 COMPARED WITH OTHER EFFICIENT NEURAL NETWORKS AFTER FINE-TUNING}
\label{tw-2}
\def\arraystretch{1}
\ignorespaces 
\centering 
\begin{tabulary}{\linewidth}{LLL}
\hline  & Top1 & Top3\\
\hline 
\multicolumn{3}{p{\dimexpr(\mcWidth{1}+\mcWidth{2}+\mcWidth{3})}}{\textbf{FOOD101}}\\
MobileNetV2 &
  74.63 &
  88.19\\

  Ensemble Net \cite{25} &
  72.12 &
  -\\
    SSGAN \cite{28} &
  75.34 &
  -\\

   MobileNetV3 &
  80.29 &
  91.01\\
\textbf{G-Specialist} &
  \textbf{80.77} &
  \textbf{91.93}\\
\multicolumn{3}{p{\dimexpr(\mcWidth{1}+\mcWidth{2}+\mcWidth{3})}}{\textbf{UECFOOD100}}\\
MobileNetV2 &
  73.72 &
  87.39\\
MobileNetV3 &
  79.50 &
  93.68\\
\textbf{G-Specialist} &
  \textbf{80.20} &
  \textbf{94.31}\\
\multicolumn{3}{p{\dimexpr(\mcWidth{1}+\mcWidth{2}+\mcWidth{3})}}{\textbf{UECFOOD256}}\\
MobileNetV2 &
  58.38 &
  77.77\\
MobileNetV3 &
  66.49 &
  84.65\\
\textbf{G-Specialist} &
  \textbf{68.00} &
  \textbf{85.00}\\
\hline 
\end{tabulary}\par 
\end{table}

\begin{table}[!htbp]
\caption{{ACCURACY (\%) ON PRISTINE DATA FOR MALAYSIAN FOOD COMPARED WITH OTHER EFFICIENT NEURAL NETWORKS AFTER FINE-TUNING} }
\label{tw-3}
\def\arraystretch{1}
\ignorespaces 
\centering 
\begin{tabulary}{\linewidth}{LLL}
\hline  & Top1 & Top5\\
\hline 
\multicolumn{3}{p{\dimexpr(\mcWidth{1}+\mcWidth{2}+\mcWidth{3})}}{\textbf{Malaysian Food}}\\
MobileNetV2 &
  60.40 &
  79.51\\
MobileNetV3 &
  68.60 &
  86.10\\
\textbf{G-Specialist} &
  \textbf{70.70} &
  \textbf{87.61}\\
\hline 
\end{tabulary}\par 
\end{table}

\section{Conclusion}
Observably, efficient neural networks trained on pristine images had poor classification performance for the distorted images affected by blur or additive noise. The G-Specialist in the study addressed the issue by modeling a generalized specialist approach to increase data value by generating synthetic images on the fly during fine-tuning, helping each model in an ensemble retain the general skill of recognizing clean images and deep expertise on all intensity levels of a particular distortion. Significantly, the study proved that each model in the ensemble individually contributed to a better prediction for noisy test images by becoming an expert on a particular noise, such as blur or additive noise.
\par  Secondly, the study presented that the models in an ensemble converge to different local minima improved classification performance on pristine data compared to a single model. Most importantly, the study has experimented on Malaysian Food database publicly available for further enrichment and evaluation besided performing experiments on three publicly available real-world image datasets. The results showed that the proposed G-Specialist consistently outperformed other methods on pristine images and recognized noisy test images with great precision than the other methodologies.



%

\bibliographystyle{IEEEtran}
\bibliography{refrences}

\bibliography{\jobname}

\begin{thebibliography}{1}
\bibitem{4} Lina J. Karam and Tong Zhu "Quality labeled faces in the wild (QLFW): a database for studying face recognition in real-world environments", Proc. SPIE 9394, Human Vision and Electronic Imaging XX, 93940B (17 March 2015); https://doi.org/10.1117/12.2080393
\bibitem{5} S. Karahan, M. Kilinc Yildirum, K. Kirtac, F. S. Rende, G. Butun and H. K. Ekenel, "How Image Degradations Affect Deep CNN-Based Face Recognition?," 2016 International Conference of the Biometrics Special Interest Group (BIOSIG), 2016, pp. 1-5, doi: 10.1109/BIOSIG.2016.7736924.
\bibitem{6} Rodner, Erik \& Simon, Marcel \& Fisher, Robert \& Denzler, Joachim. (2016). Fine-grained Recognition in the Noisy Wild: Sensitivity Analysis of Convolutional Neural Networks Approaches. 
\bibitem{9} Deep object classification in low resolution LWIR imagery via transfer learning Abbott, R., Del Rincon, J. M., Connor, B., \& Robertson, N. (2017). In Proceedings of the 5th IMA Conference on Mathematics in Defence
\bibitem{1}  Borkar, Tejas \& Karam, Lina. (2017). DeepCorrect: Correcting DNN Models Against Image Distortions. IEEE Transactions on Image Processing. PP. 10.1109/TIP.2019.2924172. 
\bibitem{2}  Dodge, Samuel \& Karam, Lina. (2016). Understanding How Image Quality Affects Deep Neural Networks. 
\bibitem{3} Gazioğlu, Bilge \& Kamasak, Mustafa. (2021). Effects of objects and image quality on melanoma classification using deep neural networks. Biomedical Signal Processing and Control. 67. 102530. 10.1016/j.bspc.2021.102530. 
\bibitem{7} Zheng, Stephan \& Song, Yang \& Leung, Thomas \& Goodfellow, Ian. (2016). Improving the Robustness of Deep Neural Networks via Stability Training. 10.1109/CVPR.2016.485. 
\bibitem{8} Z. Sun, M. Ozay, Y. Zhang, X. Liu and T. Okatani, "Feature Quantization for Defending Against Distortion of Images," 2018 IEEE/CVF Conference on Computer Vision and Pattern Recognition, 2018, pp. 7957-7966, doi: 10.1109/CVPR.2018.00830.
\bibitem{10} Huang, Gao \& Li, Yixuan \& Pleiss, Geoff \& Liu, Zhuang \& Hopcroft, John \& Weinberger, Kilian. (2017). Snapshot Ensembles: Train 1, get M for free. 
\bibitem{11} Ghorbani, Amirata \& Zou, James. (2019). Data Shapley: Equitable Valuation of Data for Machine Learning. 
\bibitem{12} Ilya Loshchilov and Frank Hutter. Sgdr: Stochastic gradient descent with restarts. arXiv preprint
arXiv:1608.03983, 2016.
\bibitem{13}Bossard, Lukas \& Guillaumin, Matthieu \& Van Gool, Luc., "Food-101 – Mining Discriminative Components with Random Forests," pp. 446-461, 2014. 
\bibitem{14}Y. Matsuda and K. Yanai, "Multiple-food recognition considering co-occurrence employing manifold ranking," in Proceedings of the 21st International Conference on Pattern Recognition (ICPR2012), Nov 2012. 

\bibitem{15} Automatic Expansion of a Food Image Dataset Leveraging Existing Categories with Domain Adaptation, "Proc. of ECCV Workshop on Transferring and Adapting Source Knowledge in Computer Vision (TASK-CV)","2014"

\bibitem{16} Dabov, Kostadin \& Foi, Alessandro \& Katkovnik, Vladimir \& Egiazarian, Karen. (2009). BM3D Image Denoising With Shape-Adaptive Principal Component Analysis. Proc. Workshop on Signal Processing with Adaptive Sparse Structured Representations (SPARS'09). 

\bibitem{17} Dong, Weisheng \& Zhang, Lei \& Shi, Guangming \& li, Xin. (2012). Nonlocally Centralized Sparse Representation for Image Restoration. IEEE transactions on image processing : a publication of the IEEE Signal Processing Society. 22. 10.1109/TIP.2012.2235847.
\bibitem{18} Howard, Andrew \& Pang, Ruoming \& Adam, Hartwig \& Le, Quoc \& Sandler, Mark \& Chen, Bo \& Wang, Weijun \& Chen, Liang-Chieh \& Tan, Mingxing \& Chu, Grace \& Vasudevan, Vijay \& Zhu, Yukun. (2019). Searching for MobileNetV3. 1314-1324. 10.1109/ICCV.2019.00140. 

\bibitem{19} Howard, Andrew \& Zhu, Menglong \& Chen, Bo \& Kalenichenko, Dmitry \& Wang, Weijun \& Weyand, Tobias \& Andreetto, Marco \& Adam, Hartwig. (2017). MobileNets: Efficient Convolutional Neural Networks for Mobile Vision Applications.

\bibitem{20} Sandler, Mark \& Howard, Andrew \& Zhu, Menglong \& Zhmoginov, Andrey \& Chen, Liang-Chieh. (2018). MobileNetV2: Inverted Residuals and Linear Bottlenecks. 4510-4520. 10.1109/CVPR.2018.00474. 
\bibitem{21} Lundberg, Scott \& Lee, Su-In. (2017). A Unified Approach to Interpreting Model Predictions. 

\bibitem{22} Ghalib Ahmed Tahir, https://github.com/ghalib2021/-G-Specialist

\bibitem{23} Patrick McAllister, Huiru Zheng, Raymond Bond, Anne Moorhead,Combining deep residual neural network features with supervised machine learning algorithms to classify diverse food image datasets,Computers in Biology and Medicine,2018
\bibitem{24} M. Niki \& P. Claudio \& M. Christian., "A supervised extreme learning committee for food recognition",Computer Vision and Image Understanding
 2016. 
 \bibitem{27} K. Yanai and Y. Kawano, "Food image recognition using deep convolutional network with pre-training and fine-tuning,” inProc. IEEE Int. Conf.Multimedia Expo. Workshops, 2015, pp. 1–6
 \bibitem{25} Pandey,A.Deepthi,B.Mandal,and N.B.Puhan,“Foodnet:Recognizing foods using ensemble of deep networks,”IEEE Signal Process. Lett.,vol. 24, no. 12, pp. 1758–1762, Dec. 2017.
 \bibitem{26} Liu, Chang et al. "A New Deep Learning-Based Food Recognition System for Dietary Assessment on An Edge Computing Service Infrastructure," IEEE Transactions on Services Computing, pp. 1-1, 2017. 
 \bibitem{28} Mandal, Bappaditya \& Puhan, Niladri \& Verma, Avijit. (2018). Deep Convolutional Generative Adversarial Network Based Food Recognition Using Partially Labeled Data. IEEE Sensors Letters. PP. 1-1. 10.1109/LSENS.2018.2886427. 
\end{thebibliography}

\end{document}